%% file: paper.tex
\documentclass[10pt,twocolumn,letterpaper]{article}

\usepackage[final]{cvpr}      

\usepackage{graphicx}
\usepackage{amsmath}
\usepackage{amssymb}
\usepackage{booktabs}

\usepackage[linesnumbered,ruled]{algorithm2e}
\usepackage{algpseudocode}

\usepackage[utf8]{inputenc} 
\usepackage[T1]{fontenc}    
\usepackage{url}            
\usepackage{amsfonts}       
\usepackage{nicefrac}       
\usepackage{microtype}      
\usepackage{xcolor}         
\usepackage{wrapfig}
\usepackage{bm}
\usepackage{bbm}
\usepackage{mathtools}
\usepackage{listings}
\usepackage{xcolor,colortbl}

\usepackage[frozencache,cachedir=.]{minted}
\usepackage{comment}
\definecolor{Gray}{gray}{0.85}
\definecolor{LightCyan}{rgb}{0.85,0.92,0.90}

\DeclareMathOperator*{\argmax}{argmax}

\def\vs{{\em vs.}}

\newcommand{\one}[1]{\mathbbm{1}_{[#1]}}

\usepackage[pagebackref,breaklinks,colorlinks]{hyperref}

\usepackage[capitalize]{cleveref}
\crefname{section}{Sec.}{Secs.}
\Crefname{section}{Section}{Sections}
\Crefname{table}{Table}{Tables}
\crefname{table}{Tab.}{Tabs.}

\def\cvprPaperID{2872} 
\def\confName{CVPR}
\def\confYear{2022}

\begin{document}

\title{Deep Bregman Divergence for \\Contrastive Learning of Visual Representations}

\author{
Mina Rezaei\textsuperscript{\rm 1}\thanks{ Equal Contributions.}, 
Farzin Soleymani\textsuperscript{\rm 2*}, 
Bernd Bischl\textsuperscript{\rm 1}, 
Shekoofeh Azizi\textsuperscript{\rm 3}\thanks{mina.rezaei@stat.uni-muenchen.de, shekazizi@google.com}\\
 \textsuperscript{\rm 1} Department of Statistics, LMU Munich, Germany\\
 \textsuperscript{\rm 2} Department of Electrical and Computer Engineering, Technical University of Munich, Germany\\
 \textsuperscript{\rm 3} Google Research, United States\\
}

\maketitle

\input{chapters/abstract.tex}
\input{chapters/introduction.tex}
\input{chapters/relatedwork.tex}
\input{chapters/method.tex}
\input{chapters/implementation.tex}
\input{chapters/experiments.tex}

\input{chapters/ablation.tex}

\input{chapters/conclusion.tex}

\vspace{-6pt}
{\small
\bibliographystyle{ieee_fullname}
\bibliography{egbib}
}

\newpage

\end{document}

%% file: chapters/abstract.tex
\vspace*{-5pt}
\begin{abstract}
\vspace{-5pt}
Deep Bregman divergence measures divergence of data points using neural networks which is beyond Euclidean distance and capable of capturing divergence over distributions. In this paper, we propose deep Bregman divergences for contrastive learning of visual representation where we aim to enhance contrastive loss used in self-supervised learning by training additional networks based on functional Bregman divergence. In contrast to the conventional contrastive learning methods which are solely based on divergences between single points, our framework can capture the divergence between distributions which improves the quality of learned representation. We show the combination of conventional contrastive loss and our proposed divergence loss outperforms baseline and most of the previous methods for self-supervised and semi-supervised learning on multiple classifications and object detection tasks and datasets. Moreover, the learned representations generalize well when transferred to the other datasets and tasks. The source code and our models are available in supplementary and will be released with paper.

\end{abstract}

%% file: chapters/introduction.tex
\vspace{-10pt}
\section{Introduction}

Metric learning aims to construct task-specific distance from supervised data, where learned distance metrics can be used to perform various tasks such as classification, clustering, and information retrieval. We can divide the extension of conventional metric learning or Mahalanobis metric learning algorithms into two directions: those methods based on deep metric learning and those based on Bregman divergence learning. Deep metric learning uses neural networks to automatically learn discriminative features from samples and then compute the metric such as contrastive loss in Siamese network~\cite{koch2015siamese} or triplet loss in the triplet network~\cite{hoffer2015deep}. Bregman divergences learning generalizes measures like Euclidean distance~\cite{bregman1967relaxation} and KL divergences~\cite{painsky2019bregman} by learning the underlying generating function of the Bregman divergence using piecewise linear approximation~\cite{siahkamari2020learning}, or by adding quantization rates to the existing basis of functional Bregman for clustering~\cite{liu2016clustering}. 

Recently, deep divergence learning~\cite{cilingir2020deep, kampffmeyer2019deep} was introduced to learn and parameterize functional Bregman divergence using linear neural networks. This method measures divergences of data by approximating functional Bregman divergences by training deep neural networks where it aims to reduce the distance between feature vectors corresponding to the same class and increase the distance between the feature vectors corresponding to different classes. A key advantage of this method is to shift the divergence learning focus from learning divergence between single points to capturing the divergence between the distribution of a vector of points~\cite{cilingir2020deep}. This can potentially introduce new desirable properties to a learned feature space. Nevertheless, learning Bregman divergence has not yet seen such widespread adoption and it remains a challenging endeavor for the representation learning and self-supervision based on distance reduction (i.e. similarity maximization).

\begin{figure} [!t]
\centering
\vspace{-5pt}
\includegraphics[width=0.46\textwidth]{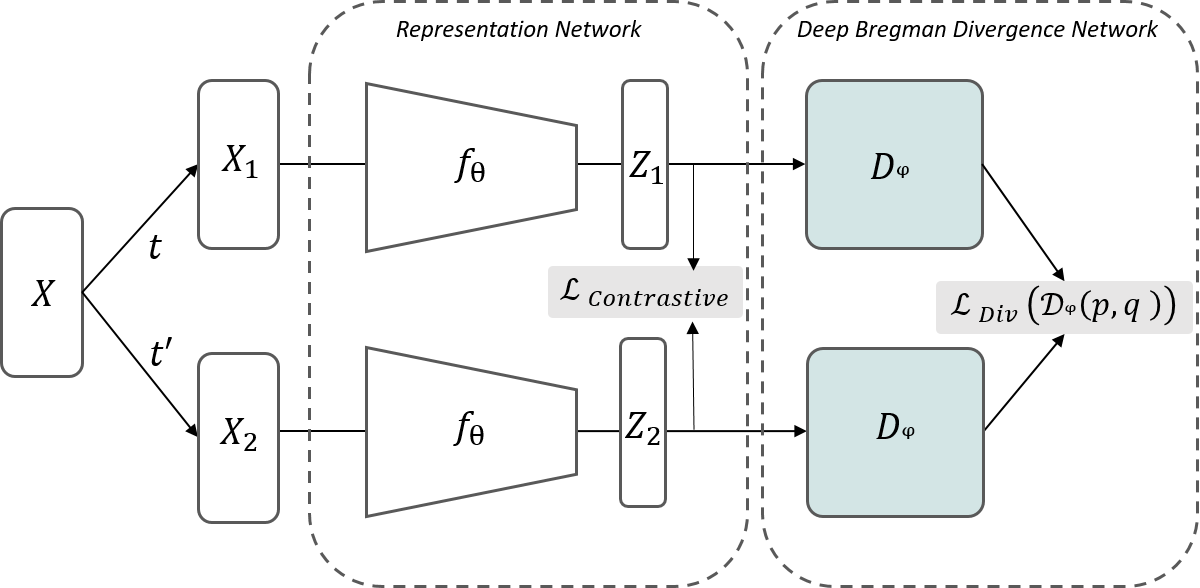}
\caption{Illustration of our proposed framework. First two augmented samples are generated by the augmentation function $t$ and $t^{\prime}$. Then augmented samples are transformed into vectors using the base encoder $f_{\theta}$. Next, they are projected onto a lower-dimensional space $z$. Then, $\kappa$ different subnetworks map the vector $z$ into arbitrary Bregman divergence. In these configurations, the underlying Bregman divergence is parameterized by a convex set of linear functions. Finally, contrastive divergence loss measures the underlying Bregman divergence of distribution for current mini-batch samples.} 
\vspace{+2pt}
\label{fig:method}
\end{figure}

Self-supervised learning is amongst the most promising approaches for learning from limited labeled data. In contrast to supervised methods, these techniques learn the representation of the data without relying on human annotation. Early self-supervised methods focused on solving a \textit{pretext task} such as predicting the rotation of images, the relative position of patches, and colorization~\cite{lee2017unsupervised,noroozi2016unsupervised,pathak2016context,zhang2016colorful}. Most recent successful self-supervised methods use contrastive loss and maximize the similarity of representation obtained from different distorted versions of a sample. Instance discrimination~\cite{wu2018unsupervised}, CPC~\cite{henaff2019data,oord2018representation}, Deep InfoMax~\cite{hjelm2018learning}, AMDIM~\cite{bachman2019learning}, CMC~\cite{stojnic2021self}, PIRL~\cite{misra2020self}, MoCo~\cite{he2020momentum}, and SimCLR~\cite{chen2020simple, chen2020big} are examples of such contrastive learning methods that produce representations that are competitive with supervised ones. However, existing deep contrastive learning methods are not directly amenable to comparing the data distributions.  
In this paper, we extend existing contrastive learning approaches in a more profound way using the \emph{contrastive divergence learning} strategy which learns a generalized divergence of the data distribution by jointly adopting a functional Bregman divergence along with a contrastive learning strategy. We train our framework end-to-end in four stages: (1) Transformation: we sequentially apply random color distortions, random rotations, random cropping followed by resizing back to the original size, and random Gaussian blur. Then, we train our network with two distorted images only. (2) Base network: learns representations on the top of distorted samples using deep neural networks. These representations are projected onto a lower-dimensional space by the projection head. We call the output of the base network the \textit{representations} and the output of the projector the \textit{embeddings}. (3) Bregman divergence network: the $\kappa$ different deep linear neural \textit{subnetworks} create the convex generating function and are able to compute divergence between two distributions given embeddings. (4) Contrastive divergence learning: Our network is trained end-to-end by combining a conventional contrastive loss and a new divergence loss. As depicted in Fig.~\ref{fig:method}, the contrastive loss computed over embeddings and our novel divergence loss formulated over the output of subnetworks. Our main findings and contributions can be summarized as follows:

\begin{itemize}
    \itemsep0em 
    \item We propose a framework for self-supervised learning of visual representations using functional Bregman divergence. Our proposed method learns distances that are beyond Euclidean and are capable of capturing divergence over distributions while also benefiting from the contrastive learning mechanism. 
    \item We propose a contrastive divergence loss which encourages each subnetwork to focus on different attributes of an input. The combination of our novel contrastive divergence loss with conventional contrastive loss (i.e. NT-Xent, NT-Logistic) improves the performance of contrastive learning for multiple tasks including image classification and object detection over baselines. 
    \item We show empirical results to highlight the benefits of learning representation using our methods and the proposed contrastive divergence loss. Our method is evaluated in linear and semi-supervised settings on public datasets, achieving sizable gain over baselines and comparable or higher performance in comparison to state-of-the-art contrastive learning methods in several tasks. Moreover, we show the learned representations generalize well when transferred to a new task.
\end{itemize} 

%% file: chapters/relatedwork.tex
\vspace{-10pt}
\section{Related Work}
\vspace{-2pt}

\paragraph{Divergence learning.}
Our proposed method is comparable with approaches involving measuring or learning divergence over distributions such as contrastive divergence~\cite{carreira2005contrastive}, stochastic Bregman divergence~\cite{dragomir2021fast}, maximum mean discrepancy~\cite{xing2002distance}, and most recently deep divergence learning~\cite{cilingir2020deep,kampffmeyer2019deep,kong2020rankmax,kato2021non}. Another related work is Rankmax~\cite{kong2020rankmax} which studies an adaptive projection alternative to the softmax function that is based on a projection on the $(n,k)-$simplex with application in multi-class classification. However, our proposed method differs in representation learning strategy and applications.

\vspace{-5pt}
\paragraph{Self-supervised learning.}
Much of the early works on self-supervised learning focused on the problem of learning embeddings without labels such that the linear classifier operating on the learned embeddings from self-supervision could achieve high classification accuracy~\cite{doersch2015unsupervised}. Later, some models aim to learn the representation using auxiliary handcrafted prediction tasks. Examples are image jigsaw puzzle~\cite{noroozi2016unsupervised}, relative patch prediction~\cite{doersch2015unsupervised,doersch2017multi}, image inpainting~\cite{pathak2016context}, image super-resolution~\cite{ledig2017photo,yuksel2021latentclr}. Contrastive learning is amongst the most successful self-supervised method to achieve linear classification accuracy and outperforming supervised learning tasks by suitable architectures and loss~\cite{caron2020unsupervised,chen2020simple,chen2020big,chen2020intriguing,zbontar2021barlow}, using pretraining in a task-agnostic fashion~\cite{kolesnikov2019revisiting,shen2020mix}, and fine-tuning on the labeled subset in a task-specific fashion~\cite{wu2018unsupervised,henaff2019data}. However, Hjelm et al.~\cite{hjelm2018learning} showed the accuracy depends on a large number of negative samples in the training batch. 

Self-supervised training with a large batch size is computationally expensive for high-resolution images. BYOL~\cite{grill2020bootstrap} and SimSiam~\cite{chen2021exploring} mitigated this issue by an additional prediction head and learning the latent representations of positive samples only. Robinson et al.~\cite{robinson2020contrastive} proposed a technique for selecting negative samples and the whitening procedure of~\cite{ermolov2021whitening} despite success is still sensitive to batch size. MoCo~\cite{he2020momentum,chen2020improved} alleviates this problem using a memory-efficient queue of the last visited negatives, together with a momentum encoder that preserves the intra-queue representation consistency. Most recently, SSL-HSIC~\cite{li2021self} proposed a method to maximize dependence between representations of transformed versions of an image and the image identity, while minimizing the kernelized variance of those features. VICReg~\cite{bardes2021vicreg} proposed a regularization term on the variance of the embeddings to explicitly avoid the collapse problem. Unlike other methods for self-supervised learning~\cite{chen2020simple,chen2020big,zbontar2021barlow,he2020momentum, grill2020bootstrap,ermolov2021whitening}, our proposed method simultaneously is trained by deep representation framework and deep divergence framework. We developed and studied the impact of $\kappa$-different parameterized convex linear neural networks with functional Bregman divergence on top of a simple contrastive learning framework~\cite{chen2020simple,chen2020improved}.

%% file: chapters/method.tex
\vspace{-0pt}
\section{Problem Formulation and Approach}
Our goals are twofold: First, we aim to learn representations using contrastive loss on top of embeddings via the representation network. Second, we learn a deep Bregman divergence by minimizing divergence between samples from the same distribution and maximizing divergence for samples from different classes and various distributions. As depicted in Fig.~\ref{fig:method}, our proposed method includes two sequentially connected neural networks: the \emph{representation network} and the \emph{deep Bregman divergence network}. In the following sections, we first describe contrastive learning in the context of visual representation. Then we discuss Bergman divergence, functional Bregman, and deep Bregman divergence network. Next, we describe our framework and the proposed loss. 
 
\begin{figure} [!t]
\centering
    \includegraphics[width=0.49\textwidth]{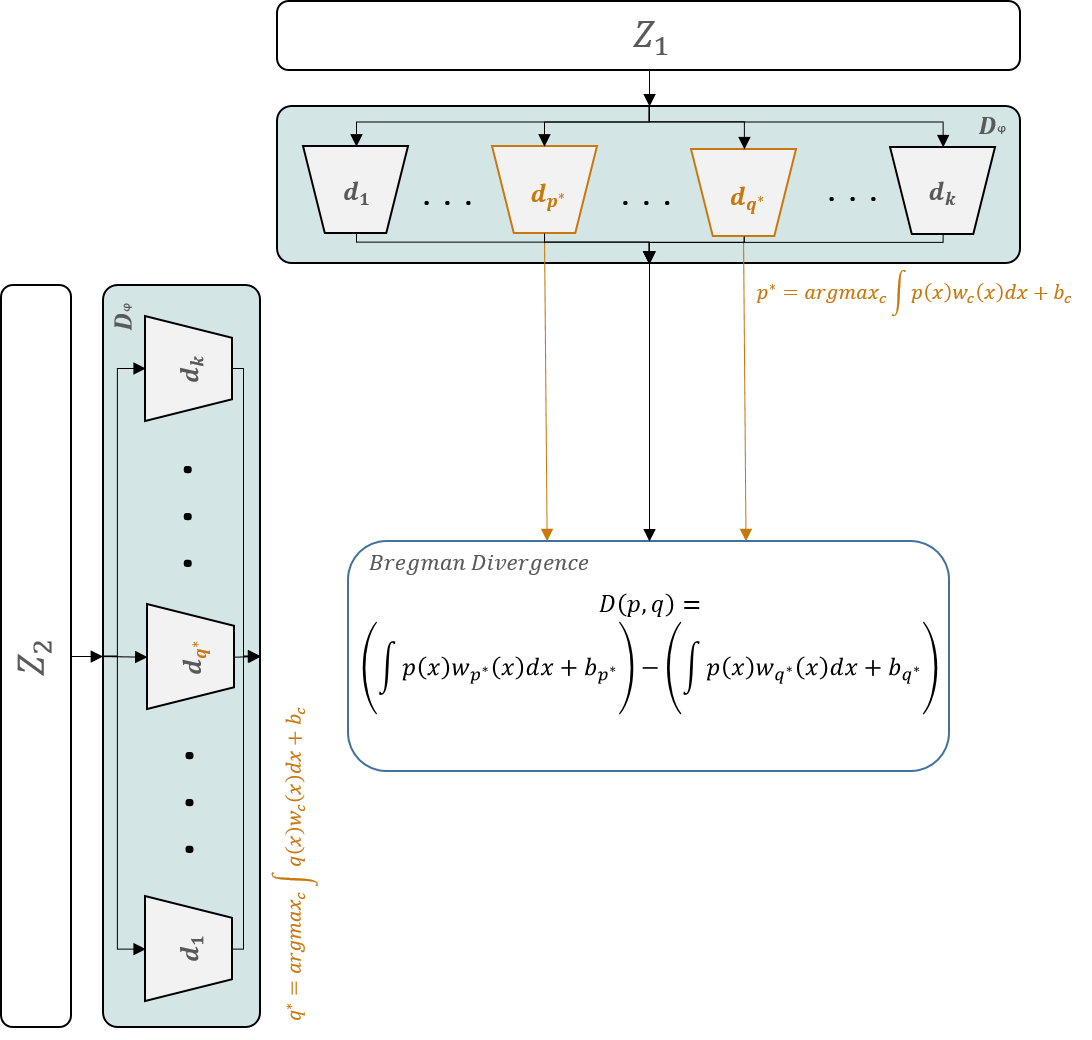}
    \caption{Illustration of the functional Bregman divergence optimization process.} 
\label{fig:architecture}
\end{figure}

\vspace{-3pt}
\subsection{Contrastive Learning} \label{sec:CLR}
Given a randomly sampled mini-batch of images $X=\{x_i\}_{i=1}^n$ with $N$ samples, contrastive learning aims to learn an embedding function $f(\cdot)$ by contrasting positive pairs $(x,x^+)$ against negative pairs $(x, x^-)$. First, we generate a positive pair sharing the same semantics from each sample in a mini-batch $X$ by applying standard image transformation techniques. For each positive pair, there exists $2(N-1)$ negative examples in a mini-batch. The encoder network (e.g. ResNet-50~\cite{he2016deep}) encodes distorted positive and negative samples to a set of corresponding features. These features are then transformed with a projection MLP head~\cite{chen2020simple} which results in $Z_1$ and $Z_2$. The contrastive estimation for a positive pair of examples $(x_i, x_j)$ is defined as: 

\vspace{-3pt}
\begin{equation}
\small
\label{eq:NTXent}
    \ell^{\mathrm{NT}\text{-}\mathrm{Xent}}_{x_i,x_j} = -log~ \frac{\exp(\mathrm{sim}(\bm z_i, \bm z_j)/\tau)}{\sum_{m=1}^{2N} \one{m \neq i}\exp(\mathrm{sim}(\bm z_i, \bm z_m)/\tau)}
\end{equation}

\noindent Where $\mathrm{sim}(\cdot,\cdot)$ is cosine similarity between two vectors, $N$ is the number of samples in a mini-batch, and $\tau$ is a temperature scalar. Loss over all the pairs formulated as:
\vspace{-3pt}
\begin{equation} \label{eq:simclr:loss}
\small
    \mathcal{L}^{\mathrm{Contrastive}}=
    \frac{1}{2N} {\sum\nolimits_{m=1}^{2N} \Bigg[\ell(2m-1,2m) + \ell(2m,2m-1)\Bigg]}
\end{equation}

\vspace{-3pt}
\subsection{Bregman Divergence Learning} \label{sec:bregman}
Bregman divergence parametrizes by a strictly convex function $\phi$ on convex set $\Omega \subseteq \mathbb{R}^d$, where $\phi$ is continuously-differentiable on relative interior of $\Omega$. The Bregman divergence associated with $\phi$ for data point $x_i, x_j \in \Omega$ calculated by: 
\vspace{-6pt}
\begin{equation} \label{eq_b}
\small
D_\phi (x_i, x_j) : \phi(x_i) - \phi(x_j) - (x_i - x_j)^T\nabla\phi(x_j)
\end{equation}

\noindent The well-known examples of Bregman divergence is the squared Euclidean distance parametrized by $\phi(x) = \frac{1}{2} \| x_i \|_2 ^2$; the KL-divergence parameterized by $\phi (x) = \sum_i x_i log x_i $; and the Itakura-Saito distance parametrized by $\phi(x) = -\sum_i log x_i$. Bregman divergences appear in various settings in machine learning and statistical learning. In optimization, Bregman et. al~\cite{bregman1967relaxation} proposed Bregman divergences as part of constrained optimization. In the unsupervised clustering, Bregman divergences provide a solution to extend the K-means algorithm beyond the convenience of the squared Euclidean distance~\cite{banerjee2005clustering}. In this paper, we use an extension of standard Bregman divergences called \textit{functional Bregman divergences}.  

A functional Bregman divergence~\cite{frigyik2008functional,ovcharov2018proper} generalizes the standard Bregman divergence for vectors and it measures the divergence between two functions or distributions. Given two functions $p$ and $q$, and a strictly convex functional $\phi$ defined on a convex set of functions which output in $\mathbb{R}$, the functional Bregman divergence formulated as:
\begin{equation} \label{eq_fb}
\small
D_{\phi} (p,q)=\phi(p) - \phi(q) -\int [p(x) - q(x)] \delta \phi(q)(x)d\mu(x)
\end{equation} 
Same as the vector Bregman divergence, the functional Bregman divergence holds several properties including convexity, non-negativity, linearity, equivalence classes, linear separation, dual divergences, and a generalized Pythagorean inequality.  Deep Bregman divergence~\cite{cilingir2020deep} parametrize the functional Bregman divergence by weight functions $w_1, w_2, ...,w_{\kappa}$ and biases $b_1, b_2, ...,b_{\kappa}$ with assumption that every generating convex functional $\phi(p)$ can be expressed in terms of linear functional. For the set of linear functional $A$, the $\phi$ defines as:

\vspace{-6pt}
\begin{equation}
   \phi(p) = \underset{(w, b_w) \in A}{sup} \int p(x)w(x)dx + b_w 
\end{equation}

\noindent and based on the underlying generating convex functional $\phi$, functional Bregman divergence can be expressed as:

\vspace{-5pt}
\begin{equation} 
\label{eq:D}
\small
D_{\phi} (p,q) = \\
\scriptsize
\bigg(\int p(x)w_{p^*}(x)dx + b_{p^*}\bigg) - \bigg(\int p(x)w_{q^*}(x)dx + b_{q^*}\bigg)
\end{equation}
\noindent with  $p$, $q$ are given by empirical distributions over input points; $p^*=\argmax_\kappa[\int p(x)w_\kappa(x)dx+b_\kappa]$; and $q^*$ is defined same as $p^*$. Therefore, we can train deep functional Bregman divergence if each of the weights and bias functions are given by $\kappa$ separate linear neural networks (Fig.~\ref{fig:architecture}).

\subsection{Contrastive Divergence Learning} \label{sec:BeCLR}

\input{chapters/algorithm}

Given a randomly sampled mini-batch of images $X$, our method takes an augmentation set $\bm T$ and draw two random augmentation $t$ and $t'$ to produces two distorted images $x_i = t(x)$ and $x_j = t'(x)$ for a single image $x$. The distorted samples are encoded via base network $f_{\theta}$ to generate corresponding representations, $z_i = f_{\theta}(x_i)$ and $z_j = f_{\theta}(x_j)$.

Next, we perform divergence learning by adopting a functional Bregman $D_{\phi}$ ( Fig.~\ref{fig:architecture}). Consider $p$ and $q$ as empirical distributions over $z_i$ and $z_j$, respectively. We parametrize our deep divergence with weight function $w_1,w_2,...,w_{\kappa}$ and biases $b_1,b_2,...,b_{\kappa}$. Each subnetwork $d=\{1,2,...,\kappa\}$ takes $z_i, z_j$ and produces a single output $w_d(z) + b_d$. Consider $p^*$ as the index of the maximum output and $q^*$ index of the maximum output across the $\kappa$~subnetworks. Now, the divergence is the difference between the output of $z_i$ at $p^*$ and the output of $z_j$ at $q^*$.

Considering each of the $\kappa$ outputs corresponds to a different class: the divergence is zero when both points achieve a maximum value for the same class, and it is non-zero otherwise. The divergence increases as the two outputs become more separated. Our method is trained with a combination of two losses; one based on discriminative features by representation network and another based on the Bregman divergence output by $\kappa$~subnetworks. 

In this paper, we estimate noise contrastive loss on top of representation vectors similar to Eq.~\ref{eq:simclr:loss}. The output of the representation network results in well-separated inter-product differences while the deep features learned by $\kappa$ different Bregman divergences result in well discriminative features with compact intra-product variance. Therefore, the combination of these is key to have better visual search engines. In addition, learning such discriminative features enables the network to generalize well on unseen images. 

We convert Bregman divergence $D$ (Eq.~\ref{eq:D}) to similarity using a Gaussian kernel $\psi_{i,j} = exp(-D/{2{\sigma}^2})$ (where $\sigma$ is adjustable parameter). All the divergences obtained with the same network are viewed as positive pairs while all other divergences obtained with a different network are considered as negative pairs. We enforce each subnetwork to have a consistent but also orthogonal effect on the feature. The divergence loss between $p(z_i)$ and $q(z_j)$ for a mini-batch of the size of $N$ define as: 

\vspace{-6pt}
\begin{equation} \label{eq:pair:div:loss}
\small
 \mathcal{\ell}_{div(p(z_i), q(z_j))}  =  -log \Big( exp (\psi_{i,j}) / \sum\nolimits_{m=1}^{N} exp (\psi_{i,m})\Big)
\end{equation}

\vspace{-6pt}
\begin{equation} \label{eq:div_loss}
\small
    \mathcal{L}^{\mathrm{divergence}}=
    \frac{1}{N} {\sum\nolimits_{m=1}^{N} \Bigg[\ell_{div}(2m-1,2m)\Bigg]}
\end{equation}

\noindent and the total loss calculated by the combination of the divergence loss (Eq.~\ref{eq:div_loss}) to the contrastive loss (Eq.~\ref{eq:simclr:loss}) which controlled by a learnable hyperparameter $\lambda$:
\begin{equation} \label{eq:total}
 \mathcal{L}_{total}  = \lambda \mathcal{L}_{contrastive} + \mathcal{L}_{divergence}
\end{equation}

At the end of training, similar to~\cite{chen2020simple} we only keep the encoder $f_{\theta}$. Algorithm~\ref{alg-1} shows PyTorch-style pseudo-code for our proposed method.

%% file: chapters/algorithm.tex
\begin{algorithm}[!t] 
\scriptsize{
\begin{minted}{python}

# model: network f + network d
# f: base network + projection mlp (Nxf_out)
# d: Bregman subnetworks
# D: Bregman divergence
# t: temperature

# load a minibatch x with N samples
for x in loader(dataset):
    x1, x2 = tau(x), tau_bar(x) # Augmentation
    z1, z2 = f(x1), f(x2) # Projection: Nxf_out
    o1, o2 = d(z1), d(z2) # Subnetworks output
    z = cat(z1, z2) # concatenation
    # part 1: calculating contrastive loss
    # cosine similarity matrix: 2Nxf_out
    sim_f = cos(z.unsqueeze(1), z.unsqueeze(0))
    pos_ab = diag(sim_f, N)
    pos_ba = diag(sim_f, -N)
    pos_f = cat(pos_ab, pos_ba) # positive logits 
    neg_f = sim_f[~pos] # negative logits
    logits_f = cat(pos_f, neg_f)
    labels_f = zeros(2*N)
    # contrastive loss
    # (Eq.2)
    loss1 = CrossEntropyLoss(logits_f/t, labels_f) 
    # part 2: calculating Bregman loss
    div_matrix = D(o1, o2) # Bregman divergence
    # converting divergence to similarity
    sim_b = div_to_sim(div_matrix)
    pos_b = diag(sim_b) # positive logits
    neg_b = sim_b[~pos_b] # negative logits
    logits_b = cat(pos_b, neg_b)
    labels_b = zeros(N)
    # Bregman loss
    loss2 = CrossEntropyLoss(logits_b, labels_b)
    # accumulation of two losses
    loss = lambda * loss1 + loss2 # (Eq.9)
    # SGD update
    loss.backward()
    update(model.params)
def D(o1, o2):
    p_star = argmax(o1)
    q_star = argmax(o2)
    # Bregman divergence (Eq.6)
    return o1[p_star] - o1[q_star]
    
\end{minted}
\vspace{-6pt}
\caption{\small{\!PyTorch-style pseudo-code for our method}}
\label{alg-1}
}
\end{algorithm}

%% file: chapters/implementation.tex
\vspace{-3pt}
\section{Experiment Setup}
\vspace{-3pt}
One advantage of our framework is learning over distributions and we do not restrict ourselves only to divergences between single points. We can also capture divergences between distributions of points similar to the maximum-mean discrepancy and the Wasserstein distance. Example applications are data generation, semi-supervised learning, unsupervised clustering, information retrieval, and Ranking. To empirically compare our proposed framework to existing contrastive models, we follow standard protocols by self-supervised learning and evaluate the learned representation by linear classification and semi-supervised tasks as well as transfer learning to different datasets and different computer vision tasks.

\vspace{-12pt}
\paragraph{Image augmentation} We define a random transformation function $\bm T$ that applies a combination of crop, horizontal flip, color jitter, and grayscale. Similar to~\cite{chen2020simple}, we perform crops with a random size from $0.2$ to $1.0$ of the original area and a random aspect ratio from $3/4$ to $4/3$ of the original aspect ratio. We also apply horizontal mirroring with a probability of $0.5$. Then, we apply grayscale with probability $0.2$, and color jittering with probability $0.8$ and with configuration $(0.4, 0.4, 0.4, 0.1)$. However, for ImageNet, we define the stronger jittering $(0.8, 0.8, 0.8, 0.2)$, crop size from $0.08$ to $1.0$, grayscale probability $0.2$, and Gaussian blurring with $0.5$ probability and $\zeta = (0.1, 2.0)$. In all the experiments, at the testing phase, we apply only resize and center crop.

\vspace{-12pt}
\paragraph{Deep representation network architecture}
Our base encoder consists of a convolutional residual network~\cite{he2016deep} with 18 layers and 50 layers with minor changes. The network is without the final classification layer instead it has two nonlinear multi-layer perceptrons (MLP) with rectified linear unit activation in between. This MLP consists of a linear layer with input size 1024 followed by batch normalization, rectified linear unit activation, and a final linear layer with output dimension 128 as embedding space. The embeddings are fed to the contrastive loss and used as an input for our deep divergence networks.

\vspace{-12pt}
\paragraph{Deep divergence network architecture (subnetworks)} 
We implemented $\kappa$-adaptive subnetworks on top of the MLP projection head. Many possible architectures are suitable to capture this type of network; we consider a simple and convex architecture where each network includes 2-layer MLPs with 128, 32, 1 hidden nodes, follow by batch normalization. We do not include activation between the layers to maintain Bregman properties and convex network. Each subnetwork own independent set of weights. We perform a Bayesian hyperparameter search to find the best number of hidden nodes and $\kappa$ subnetworks.

\vspace{-12pt}
\paragraph{Optimization}
We use the Adam optimizer~\cite{kingma2014adam} with a learning rate $0.005$, $\beta_1 = 0.5$, $\beta_2 = 0.999$, and weight decay $0.0001$ without restarts (similar to MoCo configuration). We convert Bregman divergences to similarity score using a strictly monotone decreasing function $\psi$. As explained in Method, we use Gaussian kernel and found the best $\sigma$ with Bayesian hyperparameter search for each dataset. Temperature $\tau$ sets equal to 0.1. We train our model with a mini-batch size of 512 and 2 GPUs on all small datasets, considering our primary objective is to verify the impact of our proposed method rather than to suppress state-of-the-art results. For ImageNet, we use a mini-batch size of 256 in 8 GPUs (Tesla A-100), and an initial learning rate of 0.003. We train for 400 epochs with the learning rate multiplied by 0.1 at 120 and 240 epochs which is taking around one week of training on ResNet-50. 

\vspace{-12pt}
\paragraph{Datasets and tasks}
We use the following datasets in our experiments: CIFAR 10/100~\cite{krizhevsky2009learning} are subsets of the tiny images dataset. Both datasets include 50,000 images for training and 10,000 validation images of size $32\times32$ with 10 and 100 classes, respectively. STL 10~\cite{coates2011analysis} consists of 5000 training images and 8000 test images in 10 classes with size of $96\times96$. This dataset includes 100,000 unlabeled images for unsupervised learning task. ImageNet~\cite{deng2009imagenet}, aka ILSVRC 2012, contains 1000 classes, with 1.28 million training images and 50,000 validation images. ISCI-2018~\cite{codella2019skin} is a challenge on the detection of seven different skin cancer and part of the MICCAI-2018 conference. The organizers released 10,015 dermatology scans with a size of $128 \times 128$ pixels collected from different clinics.

%% file: chapters/experiments.tex
\vspace{-3pt}
\section{Experiments and Results}


\subsection{Linear Evaluation}
One of the common evaluation protocol for self-supervised learning is freezing the base encoder $f(\cdot)$ after unsupervised pretraining and then training a supervised linear classifier on top of it. In our proposed architecture, the linear classifier is a fully connected layer followed by softmax which is connected on top of $f(\cdot)$ after removing the MLP's head and divergence network $D(\cdot)$. Our linear evaluation consists of studies on small and large datasets such as STL-10, CIFAR-10/100, ISIC-7, and ImageNet. We use the standard ResNet-50 architecture (24M parameters) in all of our evaluations. Our models are pretrained for only 400 epochs due to computational resource constraints. 

Table~\ref{table:lineareval:smalldatasets} shows the comparison of our model against the baseline under the linear evaluation on small datasets as well as the ImageNet dataset. The measured performance in top-1 accuracy suggests our method improves the baseline in small datasets significantly and achieves a sizable performance gain up to 3.3\% over the baseline on ImageNet. Table~\ref{table:lineareval:imagenet} shows ImageNet linear evaluation accuracy of our models in comparison to several recent contrastive learning methods. Our method obtains a top-1 accuracy of 72.6\% that is comparable to the state-of-the-art methods. 




\begin{table} [!t]
\small
\centering
\caption{\textbf{Baseline:} Comparison of proposed method with the baseline under linear evaluation measured by Top-1 accuracy (\%) for small datasets as well as ImageNet.} 
\vspace{-8pt}
\small
\label{table:lineareval:smalldatasets}
\begin{tabular}{l c c c c c}
\toprule
Method  & ISIC & STL & CIFAR & CIFAR  & ImageNet \\
\midrule
\rowcolor{Gray}
SimCLR      & 82.6 & 90.5 & 91.8 & 66.8 & 69.3 \\
\rowcolor{LightCyan}
Our method  & \textbf{83.8} & \textbf{92.4} & \textbf{93.2} & \textbf{69.0} & \textbf{72.6} \\
\bottomrule
\end{tabular}
\end{table}

\begin{table} [!t]
\vspace{-5pt}
\small
\centering
\caption{\textbf{Linear Evaluation on ImageNet:} Top-1 accuracy  under linear evaluation. Top three best models are \underline{underlined}. All models use a ResNet-50 encoder and prtrained for 400 epochs except models (\protect\footnotemark[1]) and (\protect\footnotemark[2]) which trained for 800 and 1000 epochs, respectively.}
\vspace{-8pt}
\small
\label{table:lineareval:imagenet}
\begin{tabular}{l c }
\toprule
Method & Top-1 Accuracy(\%) \\
\midrule
PIRL~\cite{misra2020self}           & 63.6 \\ 
CPC-v2~\cite{henaff2020data}        & 63.8 \\
CMC~\cite{caron2020unsupervised}    & 66.2 \\
\rowcolor{Gray}
SimCLR~\cite{chen2020big}   & 69.3\footnotemark[2] \\
SwAV~\cite{caron2020unsupervised} (w/o multi-crop) & 70.7 \\
SimSiam~\cite{chen2021exploring}    & 70.8 \\
MoCo-v2~\cite{chen2020improved}     & 71.1\footnotemark[1] \\
SimCLR-v2~\cite{chen2020big}        & 71.7 \\
W-MSE 4\cite{ermolov2021whitening}    & 72.5 \\
Barlow Twins~\cite{zbontar2021barlow} &\underline{73.2}\footnotemark[2] \\
BYOL~\cite{grill2020bootstrap}      &\underline{74.3\footnotemark[2]} \\
\rowcolor{LightCyan}
Our method                          &\underline{72.6}  \\
\bottomrule
\end{tabular}
\end{table}

\vspace{-2pt}
\subsection{Semi-supervised Learning} 
\vspace{-2pt}
We evaluate the performance of our models on a semi-supervised image classification task. In this task, we pretrain a standard ResNet-50 on unlabeled ImageNet examples and fine-tune a classification model using a subset of ImageNet examples with labels. We follow the semi-supervised protocol of~\cite{chen2020big} and use the same fixed splits of respectively 1\% and 10\% of ImageNet labeled training data. Table~\ref{table:semisupervised} shows the comparison of our performance against several concurrent models and the baseline SimCLR model. The result indicates using the proposed divergence mechanism we can outperform the baseline SimCLR significantly: 5.2\% when using 1\% of the data and 3.9\% when using 10\% of the data. We also have better or comparable results compared to state-of-the-arts while pretrained our network for significantly less amount of time (400 \vs 1000 epochs). 


\begin{table} [!t]
\vspace{-5pt}
\small
\centering
\caption{\textbf{Semi-supervised Learning on ImageNet}: top-1 accuracy for semi-supervised ImageNet classification using 1\% and 10\% training examples. Our model (\protect\footnotemark[2]) pretrained for 400 epochs while all of the other models (\protect\footnotemark[1]) are pretrained for 1000 epochs. }
\vspace{-8pt}
\small
\label{table:semisupervised}
\begin{tabular}{ l l l l }
\toprule
Method & 1\% & 10\%  \\
\midrule
Supervised   &25.4 &56.4  \\
\midrule
\rowcolor{Gray}
SimCLR~\cite{chen2020big}\footnotemark[1]       & 48.3 & 65.6 \\
MoCo-v2~\cite{chen2020improved}\footnotemark[1]   & 49.1 & 66.4 \\
BYOL~\cite{grill2020bootstrap}\footnotemark[1]    & 53.2 & 68.8 \\
SWAV~\cite{caron2020unsupervised}\footnotemark[1]     & 53.9 & 70.2 \\
Barlow Twins~\cite{zbontar2021barlow}\footnotemark[1]    & 55.0 & 69.7 \\
\rowcolor{LightCyan}
Our method\footnotemark[2] & 53.5 & 69.5\\
\bottomrule
\end{tabular}
\end{table}

\begin{table}[!t]
\vspace{-5pt}
\centering
\small
\caption{\textbf{Transfer Learning for Object Detection:} Transfer learning results from ImageNet with standard ResNet-50 architecture to the object detection task on VOC07+12. Results of methods indicated with (\protect\footnotemark[1]) are taken from~\cite{zbontar2021barlow}.}
\vspace{-8pt}
\small
\begin{tabular}{l l l l }
\toprule
  & \multicolumn{3}{c}{VOC07+12 detection} \\
Method & $AP_{all}$ & $AP_{50}$ & $AP_{75}$ \\
\midrule
Barlow Twins~\cite{zbontar2021barlow}\footnotemark[1] & 56.8 & 82.6 & 63.4 \\
MoCo-v2~\cite{chen2020improved}\footnotemark[1] & 57.4 & 82.5 & 64.0  \\
SimSiam~\cite{chen2021exploring}\footnotemark[1] & 57.0 & 82.4 & 63.7  \\
SwAV~\cite{caron2020unsupervised}\footnotemark[1] & 56.1 & 82.6 & 62.7  \\
\rowcolor{LightCyan}
Our method & 57.8 & 82.7 & 64.1 \\
\bottomrule
\end{tabular}
\label{table:transferlearning}
\end{table}

\vspace{-2pt}
\subsection{Transfer to Other Tasks }
\vspace{-2pt}
We further assess the generalization capacity of the learned representation on object detection. We train a Faster R-CNN~\cite{faster2015towards} model on Pascal VOC 2007 and Pascal VOC 2012 and evaluate on the test set of Pascal VOC. Table~\ref{table:transferlearning} provides a comparison of transfer learning performance of our self-supervised approach for the task of object detection. We use pre-trained ResNet-50 models on ImageNet and perform object detection on Pascal VOC07+12 dataset~\cite{everingham2010pascal}. Our results in Table~\ref{table:transferlearning} indicate that we performs comparably or better than state-of-the-art representation learning methods for this detection task.

%% file: chapters/ablation.tex
\begin{table*} [!t]
\vspace{-5pt}
\centering
\caption{Top-1 accuracy (in \%) under linear evaluation when our method is trained on top of different numbers of subnetworks and using CIFAR-10 ($\kappa$) data. The performance of the divergence learning framework depends on the representation feature space while our method significantly outperforms deep divergence.}
\vspace{-5pt}
\small
\label{table:lineareval:k}
\begin{tabular}{l l l l l l l l }
\toprule
Number of subnetworks $\bm k$ & 5 & 20 & 50 & 100 & 200 & 500 & 1000 \\
\midrule
Deep Divergence~\cite{cilingir2020deep} & 71.9 & 77.8 & 79.4 & 80.0 & 77.4 & 74.1 & 70.8 \\
\rowcolor{LightCyan}
Our Method                              & 85.2 & 88.9 & 91.0 & 92.6 & 92.8 & 89.2 & 85.5 \\
\bottomrule
\end{tabular}
\end{table*}

\subsection{Ablation Studies and Discussions}
\label{ablation}
To build intuition around the behavior and the observed performance of the proposed method, we further investigate the following aspects of our approach in multiple ablation studies: (1) the number of $\kappa$ subnetworks, (2) robustness of our algorithm in invariance to augmentations and transformations, (3) analysis of contrastive divergence loss and hyperparameters, and (4) analysis the impact of the Bregman divergence network on the quality of representation.



\begin{figure} [!t]
\centering
\includegraphics[width=0.3\textwidth]{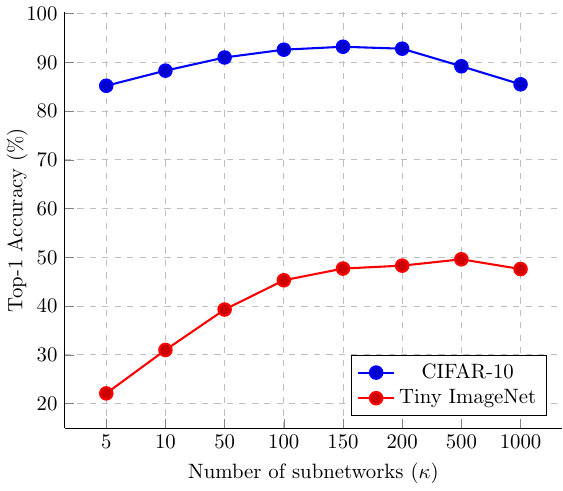}
\vspace{-6pt}
\caption{Impact of increasing subnetworks ($\kappa$) on accuracy (\%): The performance improves in both CIFAR-10 and Tiny ImageNet datasets by increasing ($\kappa$)  until a certain point, then it starts dropping possibly due to over parameterization of deep Bregman divergence subnetworks.} \label{fig:ksubnets_ablation}
\end{figure}

\noindent\textbf{Number of $\kappa$-subnetworks} We trained $\kappa$ individual deep neural networks on top of the embedding space. The input of each network was similar but they parameterized with different weights and biases. Here, we provide more details regarding our classification experiments by considering different $\kappa$. Fig.~\ref{fig:ksubnets_ablation} compares the performance in term of top-1 accuracy for CIFAR-10 and Tiny ImageNet~\cite{le2015tiny}. Based on quantitative results shown in Fig.~\ref{fig:ksubnets_ablation}, the performance improves in both CIFAR-10 and Tiny ImageNet datasets by increasing the number of subnetworks ($\kappa$) until a certain point, then it starts dropping possibly due to over parameterization. 
With a small $\kappa$, the performance of our network is more similar to contrastive loss. For example in case of CIFAR-10 shown in Fig.~\ref{fig:ksubnets_ablation}, when $\kappa = 10$ our performance is around 90\% and the performance is increased to 93.2\% for larger $\kappa$ ($\kappa = 150$). This shows training our network with a correct number of $\kappa$ can lead to a better representation of data. Table~\ref{table:lineareval:k} compares the performance of our method with Cilingir et al.~\cite{cilingir2020deep}. Based on reported results in Table~\ref{table:lineareval:k}, the performance of the divergence learning framework depends on the representation feature space while our method significantly outperforms Deep Divergence ~\cite{cilingir2020deep}. \\

\vspace{-6pt}

\vspace{-5pt}
\begin{figure} [!t]
\centering
\includegraphics[width=0.3\textwidth]{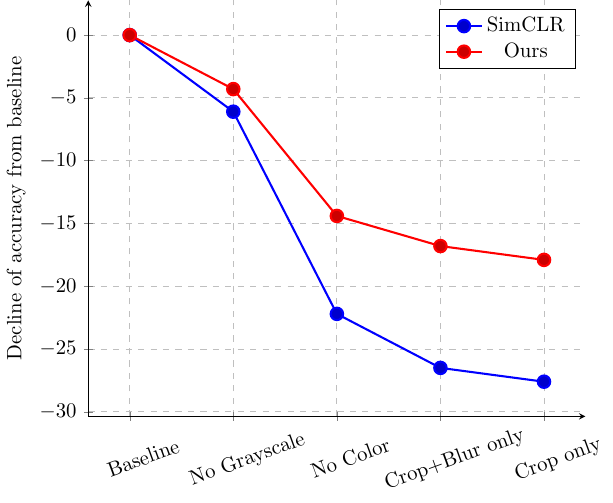}
\vspace{-6pt}
\caption{Impact of progressively removing augmentations measured by the decrease in top-1 accuracy (\%) under linear evaluation on ImageNet: our method is more robust to the effect of removing augmentation in comparison to traditional contrastive methods.} \label{fig:transformations}
\end{figure}


\noindent\textbf{Image augmentations} Since contrastive loss is sensitive to the choice of augmentation technique and learned representations can get controlled by the specific set of distortions~\cite{grill2020bootstrap}, we also examined how robust our method is to remove some of the data augmentations. Figure~\ref{fig:transformations} presents decrease in top-1 accuracy (in \% points) of our method and SimCLR and under linear evaluation on ImageNet (SimCLR numbers are extracted from~\cite{grill2020bootstrap}). This figure shows that the representations learned by our proposed contrastive divergence are more robust to removing certain augmentations in comparison to the baseline. SimCLR does not work well when removing image crop from its transformation set.

\begin{figure} [!t]
\includegraphics[width=0.49\textwidth]{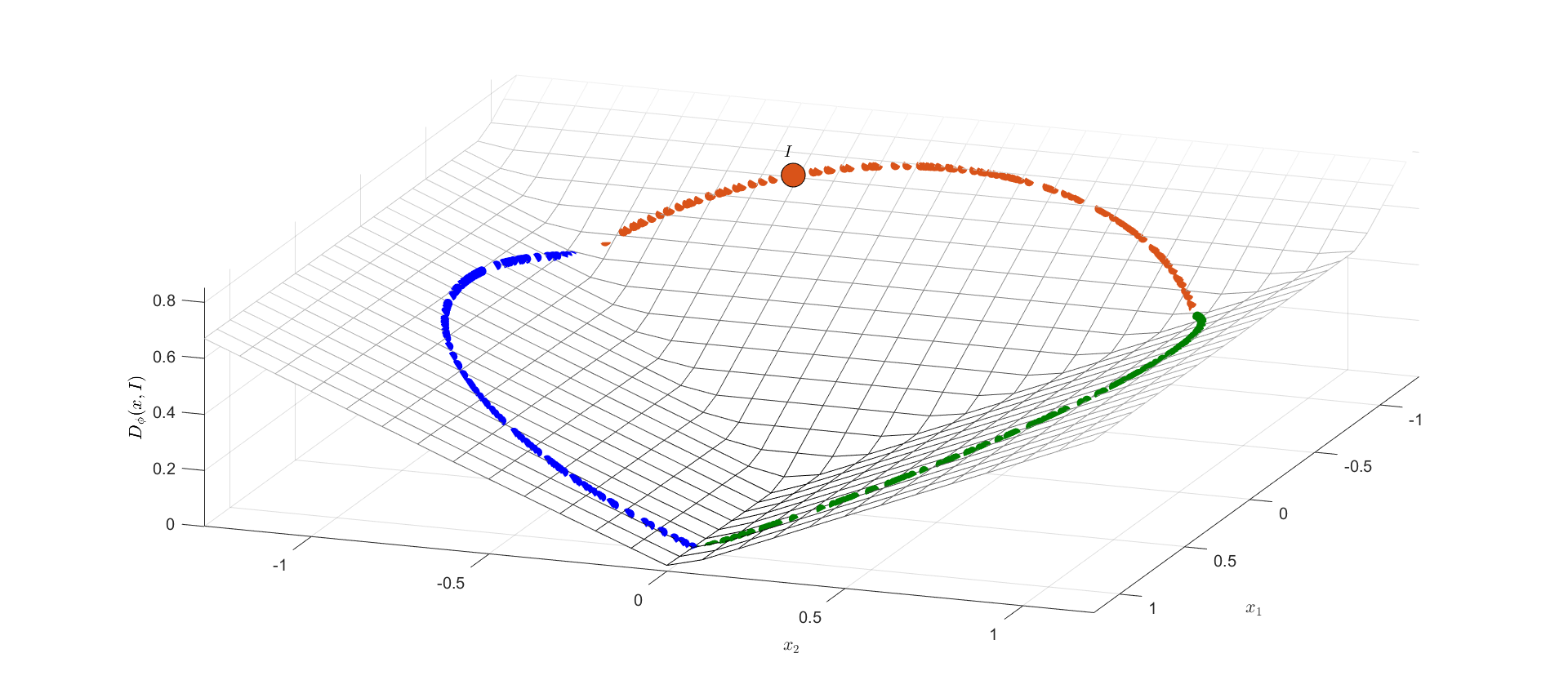}
\vspace{-6pt}
\caption{Visualization of the learned Bregman divergence on synthetic 2-D samples. We approximate the Bregman divergence from every 2-D sample point to the specific point \textit{I} in the data.} \label{fig:BD}
\end{figure}

\begin{figure}[!t]
  \centering
  \subfloat[]{\includegraphics[width=0.25\textwidth]{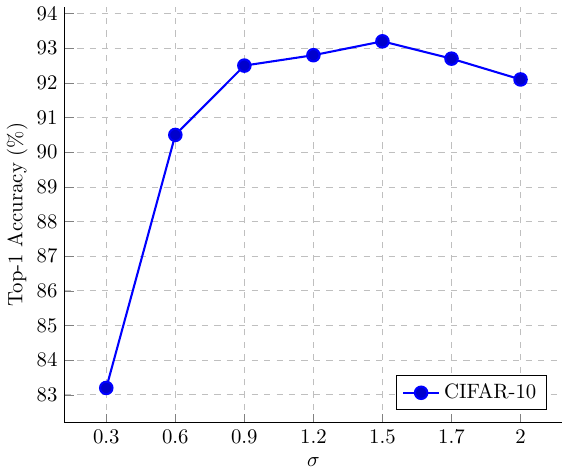}\label{fig:sigma}}
  \hfill
  \subfloat[]{\includegraphics[width=0.25\textwidth]{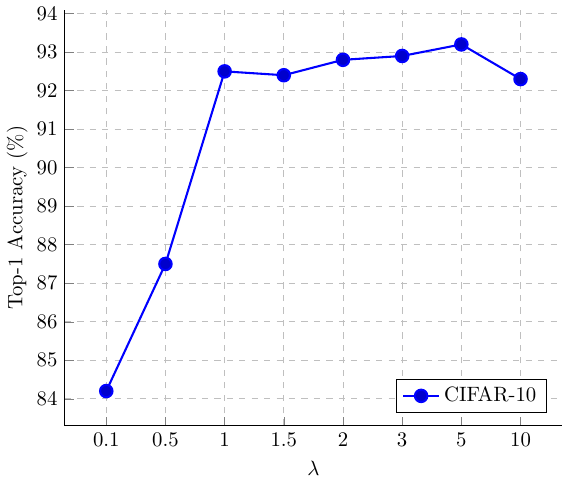}\label{fig:lamda}}
  \vspace{-6pt}
  \caption{Study of hyperparameters of contrastive divergence loss (a) $\sigma$, (b) $\lambda$ obtained by Bayesian optimization on CIFAR-10 dataset ($y$-axes).}
\label{fig:study:loss}
\end{figure}

\noindent\textbf{Impact of contrastive divergence loss} As it has been mentioned previously, the divergence loss between $p(z_i)$ and $q(z_j)$ calculated by Eq.~\ref{eq:pair:div:loss} where we convert Bregman divergence $D$ to similarity using $\psi_{i,j}$. The conversion function $\psi_{i,j}$, must be defined as a strictly monotone function and can have many forms. In general, there are multiple options to define a $\psi$ function to convert divergences to similarity scores ranging from a simple inverse function to a more complex Gaussian kernel. Table~\ref{table:ditsnceTOsimilarity} shows the functions examined in our experiments where we achieved the best performance using the Gaussian kernel. Our chosen conversion function, $\psi = exp(-D/{2{\sigma}^2})$ tuned using Bayesian optimization to find the best $\sigma$ value. We performed Bayesian hyperparameter optimization~\cite{snoek2012practical}, in order to find the best hyperparameter of $\lambda$ and $\sigma$. As depicted in Fig.~\ref{fig:study:loss}, the best Top-1 accuracy achieved when the $\lambda$ and $\sigma$ set to 5 and 1.5, respectively. 

\begin{table}[!t]
\small
\centering
\caption{Converting functional Bregman divergences ($D$) to similarity ($\psi$). There are multiple options to define a function to convert divergences to similarity scores ranging from a simple inverse function to a more complex Gaussian kernel. We achieve the best results using a Gaussian Kernel with learnable term $\sigma = 1.5$.}
\vspace{-5pt}
\small
\begin{tabular}{c} 
\toprule
 Strictly monotone function $\psi$ \\
\midrule
  $\psi = \frac{1}{1 + D} $ \\
  \rule{0pt}{3.5ex} 
  $\psi = \sqrt{(1-D)}$   \\
  \rule{0pt}{3ex} 
  $\psi = 1 - D/max(D)$  \\
  \rule{0pt}{3ex} 
  $\psi =  arccot(\alpha D)  (\alpha > 0)$  \\
\midrule
    \rule{0pt}{1ex}
    \textbf{$\psi = exp(-D/{2{\sigma}^2})$} \\
\bottomrule
\end{tabular}
\label{table:ditsnceTOsimilarity}
\end{table}

Figure~\ref{fig:BD} demonstrated approximating divergences between different classes. In this example, we show that the divergence of a random sample $(x)$ to a specific sample $I$ in the data is equal to zero when they are from the same category. In general, the outputs of the underlying convex generating functional of the input samples that are from the same category will lie on a specific hyper-plane. We use the same strategy as~\cite{siahkamari2019learning} to produce the graph.

\begin{figure}
  \centering
  \subfloat[]{\includegraphics[width=0.35\textwidth]{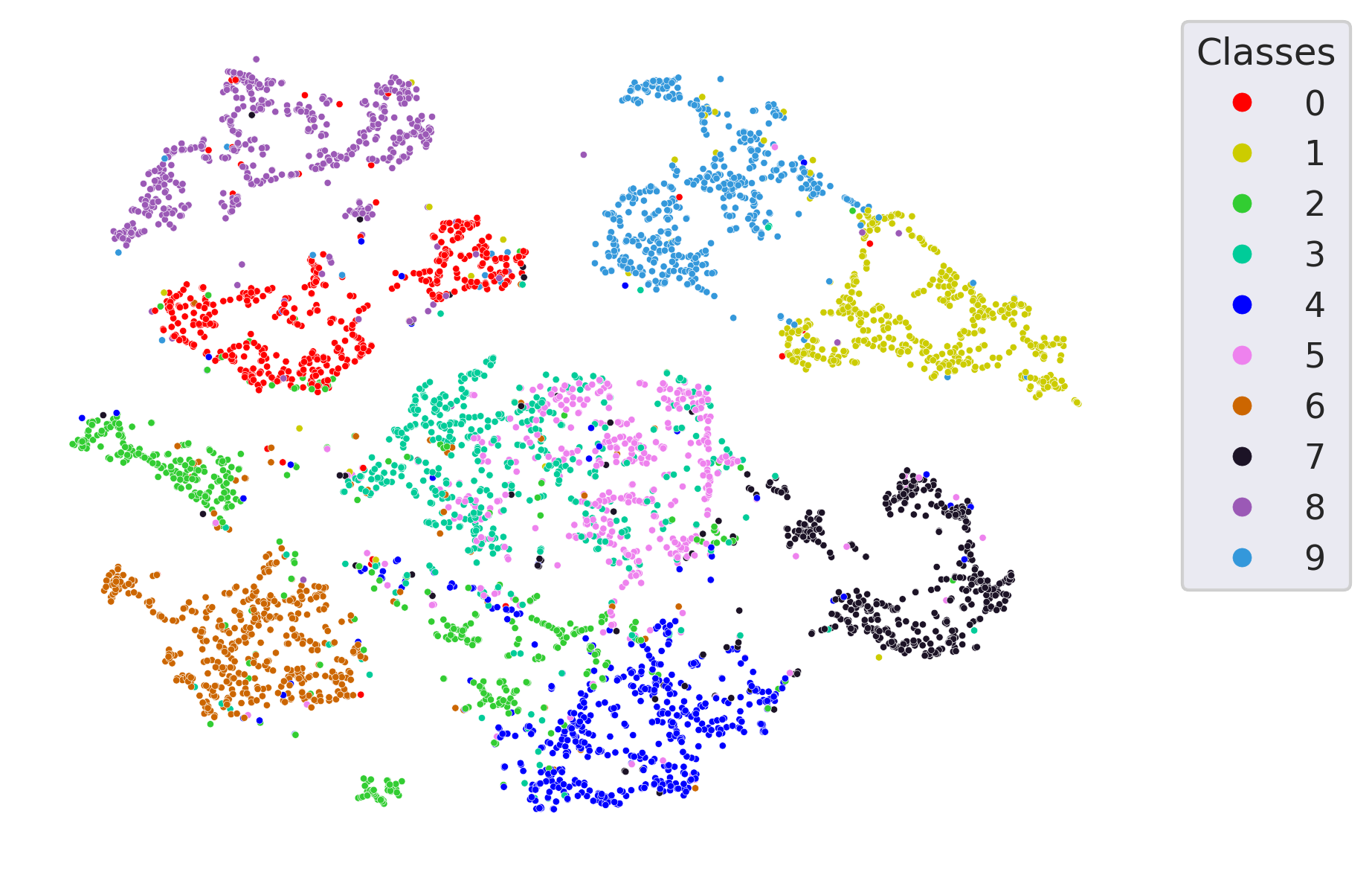}\label{fig_1a}}
  \hfill
  \subfloat[]{\includegraphics[width=0.35\textwidth]{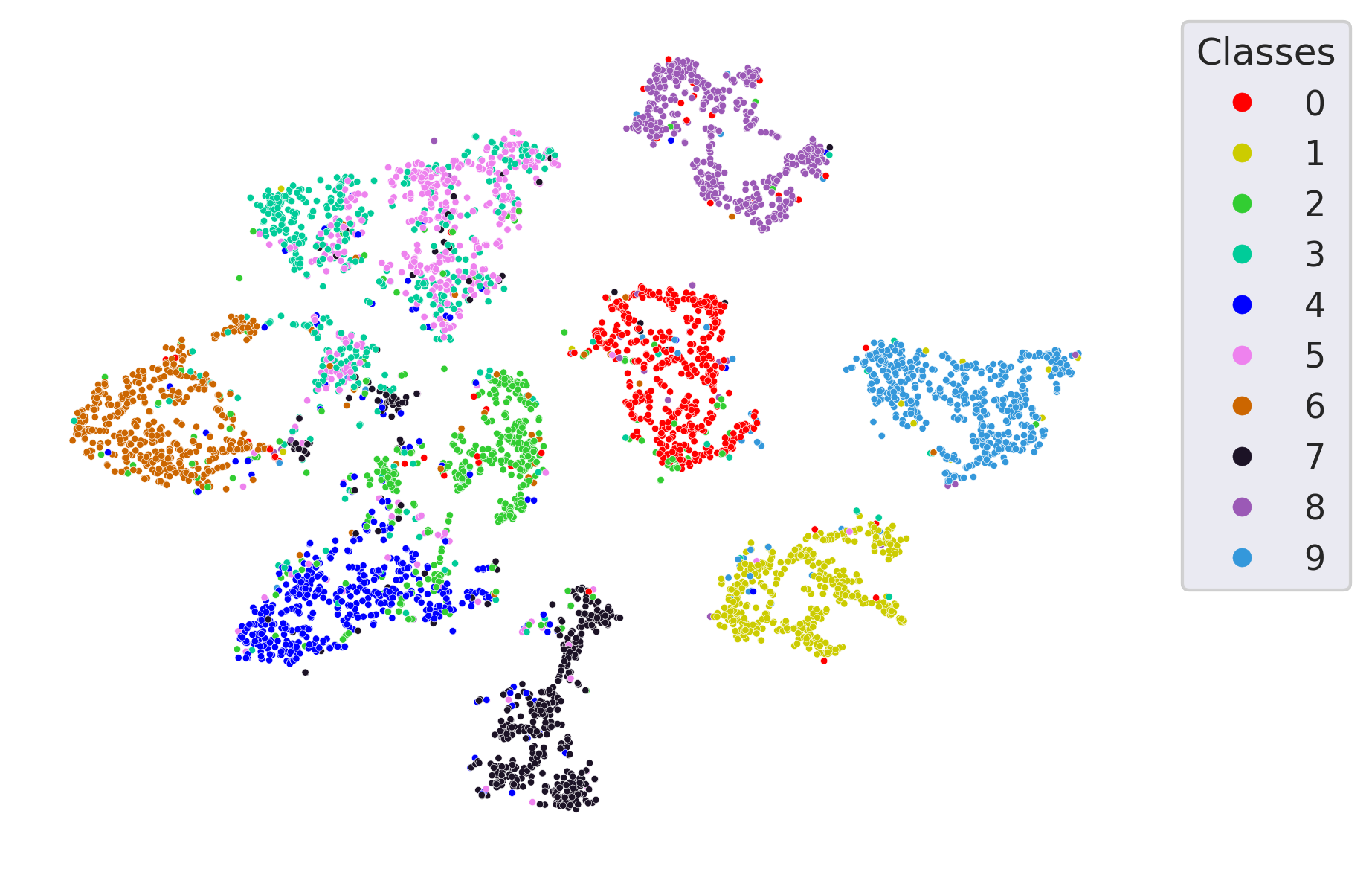}\label{fig_2b}}
  \vspace{-6pt}
  \caption{t-SNE visualization of embeddings with ResNet-50 on CIFAR-10: (a) SimCLR, (b) our method. Classes are well-clustered using our method especially for classes 0, 1, 8, and 9. }
\label{fig:tsne}
\end{figure}

\noindent\textbf{Quality of representations} We visualize the representation features using t-SNE with the last convolution layer of the ResNet-50 pretrained on CIFAR-10  to explore the quality of learned features using our proposed method. Figure~\ref{fig:tsne} compares the representation space learned by SimCLR (a) and our proposed method (b). As depicted in Fig.~\ref{fig:tsne}, our model shows better separation on clusters, especially for classes 0, 1, 8, and 9.

\subsection{Broader Impact and Limitations}
Metric learning and divergence learning are the fundamental problems in machine learning, attracting considerable research and applications. Broader impact of our work are specially on these applications include (but are not limited to) uncertainty quantification, density estimation, image retrieval, unsupervised image clustering, program debugging, image generation, music analysis, and ranking. 

One limitation of our model compared to other learning methods such as supervised learning is self-supervised learning can demand more computing resources and training time. This is somehow acceptable considering the fact that our proposed method does not need manual annotation which is usually very expensive. Furthermore, despite limited evaluation of the method on medical data, the benefit of the method in real-world applications and datasets such as robotics and medical image analysis is yet to be investigated. Also, in our study, due to computation limits and possible environmental impact of long training we limit our evaluations to a single run and eliminate error bounds obtained from several runs.  




%% file: chapters/conclusion.tex
\vspace{-8pt}
\section{Conclusion} \label{conclusion}
\vspace{-6pt}
In this paper, we proposed and examined deep divergence for contrastive learning of visual representation. Our framework is composed of the representation learning network followed by multiple divergence learning networks. We train functional Bregman divergence on top of the representation network using $\kappa$-adaptive convex neural networks. The similarity matrix is formulated according to Bregman distance output by an ensemble of the networks. Then networks are optimized end-to-end using our novel contrastive divergence loss. We successfully improve over previous methods for deep metric learning, deep divergence learning, self-supervised, semi-supervised, and transfer learning. Empirical experiments demonstrate the efficacy of the proposed method on standard benchmarks as well as recent clinical datasets on both classification and object detection tasks. 